\documentclass[10pt,twocolumn,letterpaper]{article}

\usepackage{cvpr}              

\usepackage{graphicx}
\usepackage{amsmath}
\usepackage{amssymb}
\usepackage{booktabs}
\usepackage{multirow} 
\usepackage{array}
\usepackage[pagebackref,breaklinks,colorlinks]{hyperref}

\usepackage[capitalize]{cleveref}
\crefname{section}{Sec.}{Secs.}
\Crefname{section}{Section}{Sections}
\Crefname{table}{Table}{Tables}
\crefname{table}{Tab.}{Tabs.}

\begin{document}

\title{Facial Demorphing from a Single Morph Using a Latent Conditional GAN}

\author{Nitish Shukla and Arun Ross\\
Michigan State University\\
{\tt\small {\{shuklan3,rossarun\}}@msu.edu}
}

\twocolumn[{%
\renewcommand\twocolumn[1][]{#1}%
\maketitle
\begin{center}
    \centering
    \captionsetup{type=figure}
    \includegraphics[width=.8\textwidth,height=5cm]{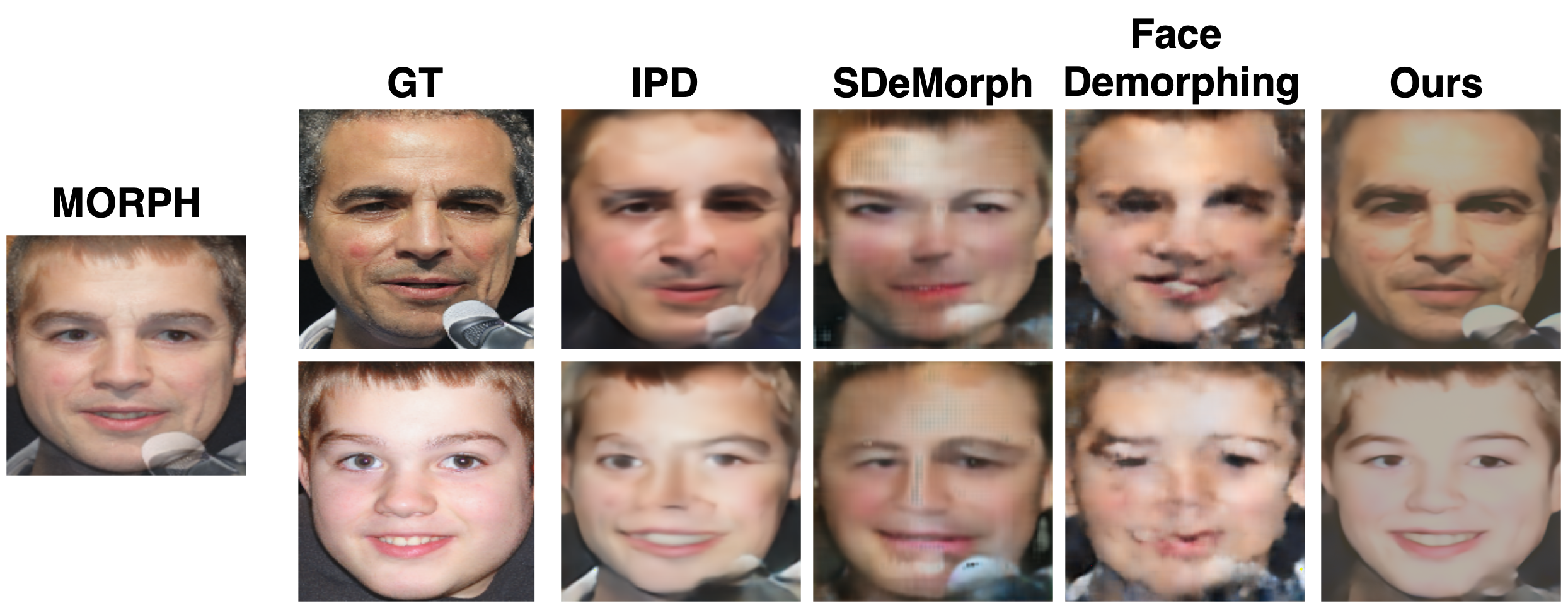}
    
    \captionof{figure}{Comparison with  current state-of-the-art demorphing methods under a unified training protocol: Our method surpasses  current methods, viz., Identity-Preserving Demorphing (IPD) \cite{ref66},  SDeMorph \cite{ref18}  and Face Demorphing \cite{ref51}, in terms of both visual quality and analytical performance. Notably, i) the outputs from our method look dissimilar from each other, and ii) outputs faithfully match the corresponding ground truth both in the pixel and latent spaces.  }
    \label{comparison}
\end{center}%
}]



\begin{abstract} A morph is created by combining two (or more) face images from two (or more) identities to create a composite image that is highly similar to all constituent identities, allowing the forged morph to be biometrically associated with more than one individual. Morph Attack Detection (MAD) can be used to detect a morph, but does not reveal the constituent images. Demorphing - the process of deducing the constituent images - is thus vital to provide additional evidence about a morph. Existing demorphing methods suffer from the morph replication problem, where the outputs tend to look very similar to the morph itself, or assume that train and test morphs are generated using the same morph technique. The proposed method overcomes these issues. The method decomposes a morph in latent space allowing it to demorph images created from unseen morph techniques and face styles. We train our method on morphs created from synthetic faces and test on morphs created from real faces using different morph techniques. Our method outperforms existing  methods by a considerable margin and produces high fidelity  demorphed face images.
\end{abstract}


\section{Introduction}
\label{sec:intro}
\textit{How can we efficiently recover the constituent images from a single face morph image?} A face morph is created by mixing two (possibly more) face images belonging to different individuals while ensuring that it biometrically matches with all participating identities \cite{ref109,ref110}. Morphs can evade manual detection \cite{10.1007/978-3-642-41181-6_75, ref11}, enabling multiple individuals to gain access using a single document \cite{ref11,ref10}, making them an obvious security threat \cite{ref12}.  Deducing constituent face images from a single morph - known as demorphing - has been a crucial yet challenging task. Demorphing is an ill-posed inverse problem, where the challenge lies not only in the absence of the knowledge of the morphing technique used (landmark-based or deep learning based) but also in the lack of constraints on the output space.  In this paper, we propose a new framework to separate constituent face images from a single morph image. Our method operates under a more realistic evaluation  (see Section \ref{background}) and achieves high facial fidelity in demorphing images (see Figure \ref{comparison}). 

Demorphing, just like morph attack detection (MAD), can be either i) reference-based \cite{ref16,ref17,ref111,ref112,ref114,9484355}, i.e., an image of one of the constituent identities is available besides the morph image, or ii) reference-free \cite{ref18,ref51,ref66}, i.e., no additional information is available besides the morph image. Single image reference-free demorphing is a significantly challenging task compared to its reference-based counterpart. Indeed, given a single morph image, with no additional constraints in output image space, there can be an infinite number of possible decomposed pairs. Moreover, developing a unified framework to demorph images created using different morphing techniques is equally challenging. Our method learns to produce the most likely set of face images conditioned on the input morph.

\textbf{Demorphing in Latent Space}: 
Learning the process of demorphing can be divided into two stages: perceptual compression, which discards unimportant details while preserving semantics, followed by generative modeling of the decomposition process in the latent space. Operating in latent space simplifies learning by offering a perceptually equivalent but more tractable domain  \cite{ref75}.
To achieve this, we first learn an autoencoder that projects the input image into a lower dimensional but perceptually equivalent and computationally efficient latent space, and second, we train a conditional GAN \cite{ref115}, conditioned on the input morph image to effectively separate it into its constituent images in this latent space.

In addition to being computationally efficient, a key advantage of our method is that, unlike prior approaches, the demorpher needs to be trained only once. This is because it operates directly on latent representations, which are already disentangled from image-level noise and artifacts. The compression stage  captures only the important semantic features, ignoring the unimportant attributes in the pixel space (background, morphing artifacts, etc.). This means that our method can handle morphs made using  unseen morphing techniques and faces despite never being trained on them. We refer to our method as \textit{Latent Conditional GAN}.

Another key challenge is the limited availability of datasets for demorphing, and privacy issues related to the use or sharing of real face data. Existing morph datasets are mainly designed for morph attack detection and typically contain only around 1.5K morphs, which is insufficient for training generative models. To address these limitations, we train our method on  morphs created using synthetically generated face images, thereby simultaneously addressing both privacy and data scarcity.  

In summary, our contributions are as follows:
\begin{itemize}
    \item In contrast to previous work \cite{ref51,ref66,ref18}, our method  has fewer constraints  and operates under a realistic protocol. 
    \item Unlike previous methods, our method is agnostic to morphing technique and face style (passport style, background, etc). Our method can be used for general demorphing (see Section \ref{sec:human}).
    \item We experimentally show that by demorphing in latent space, our method not only overcomes \textit{morph replication}, where a model tends to reproduce the morph image as its outputs, it also suppresses high-frequency artifacts in recovered images.
\end{itemize}
\begin{figure*}[]
    \centering
    \includegraphics[width=0.6\linewidth]{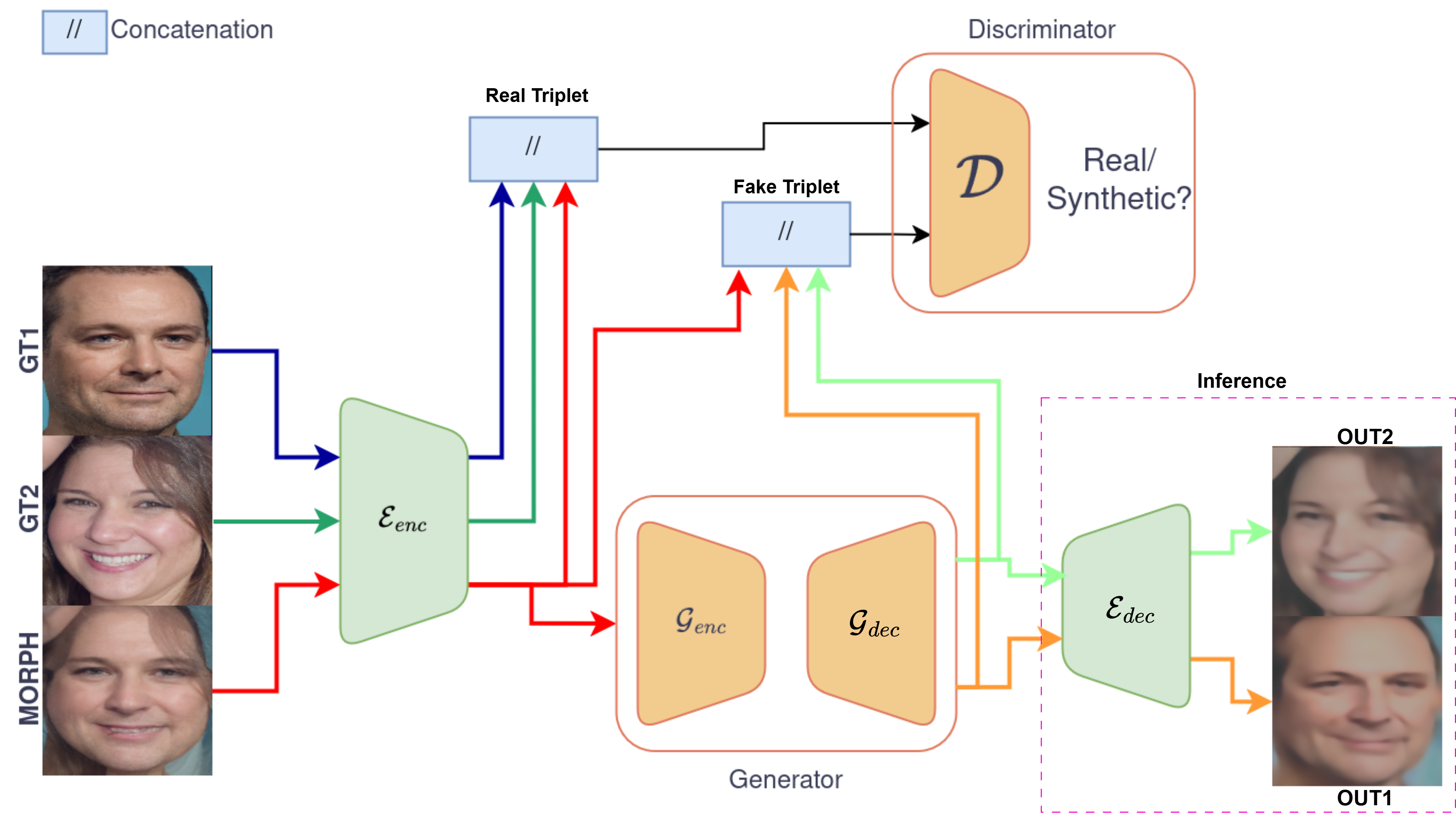}
    \caption{Proposed Demorphing Architecture: An encoder, $\mathcal{E}_{enc}$, compresses the morph along with the constituent face images during training. The generator, $\mathcal{G}$, reconstructs two face images conditioned on the morph in $\mathcal{E}_{enc}$'s latent domain. The discriminator distinguishes between real and synthesized face feature triplets. During inference, a decompressor, $\mathcal{E}_{dec}$, recovers the constituent images. Note that the decoder, $\mathcal{E}_{dec}$, is used only during inference to decompress the demorphed outputs.  }
    \label{fig:architecture}
\end{figure*}
\section{Background}
\label{background}
Morphing typically combines two face images  of two distinct individuals. Traditionally, morphing has been  accomplished by geometrically aligning facial landmarks and superimposing the second image on top of the \textit{base image} \cite{ ref68, ref70}.  Due to this formulation, morphs typically are biased toward one of the participating images, leaving very little visual information about the second image for faithful recovery. Recently, deep morphing methods have been proposed \cite{ref9,ref34,ref107,ref69} that create morphs from the ground up, making demorphing even more challenging. A morphing operator, $\mathcal{M}$, acts on two face images, $i_1$ and $i_2$, to produce the morph $x$, i.e., $x=\mathcal{M}(i_1,i_2)$. The goal of the morphing operator is to create a morph such that i) the morph looks reasonable in terms of perceptual image quality in pixel space, and ii) the morph matches both $i_1$ and $i_2$  with respect to a biometric matcher $\mathcal{B}$, i.e., $\mathcal{B}(x,i_n)>\tau$, $n\in\{1,2\}$ and $\tau$ is the matching threshold depending on the application scenario.  Reference-free demorphing attempts to approximate the inverse of the morphing process. Given a morphed input, $x$, the goal is to recover the images used to create the morph. Reference-free demorphing is an ill-posed inverse problem \cite{ref51,ref18}.

A demorphing operator, $\mathcal{DM} (=\mathcal{M}^{-1})$, upon receiving the morph image, $x$, attempts to reconstruct the original constituent images, $o_1,o_2=\mathcal{DM}(x)$,
satisfying the conditions:
\begin{equation}
\label{eq3}
\mathcal{B}(o_1,o_2)<\theta
\end{equation}
\begin{equation}
\label{eq4}
    \min_{j\in\{1,2\}} \max_{\substack{k \in \{1,2\} \\ k \neq j}} \{\ \ \mathcal{B}(o_j,i_k),\mathcal{B}(o_j,i_j)\ \ \} >\epsilon
\end{equation}
where, $\theta$ and  $\epsilon$ are matching thresholds.
Eq. (\ref{eq3}) enforces the reconstructed outputs to appear dissimilar to each other and Eq. (\ref{eq4}) ensures that each reconstructed output aligns with its corresponding ground truth image.

\begin{figure}
    \centering
    \includegraphics[width=\linewidth]{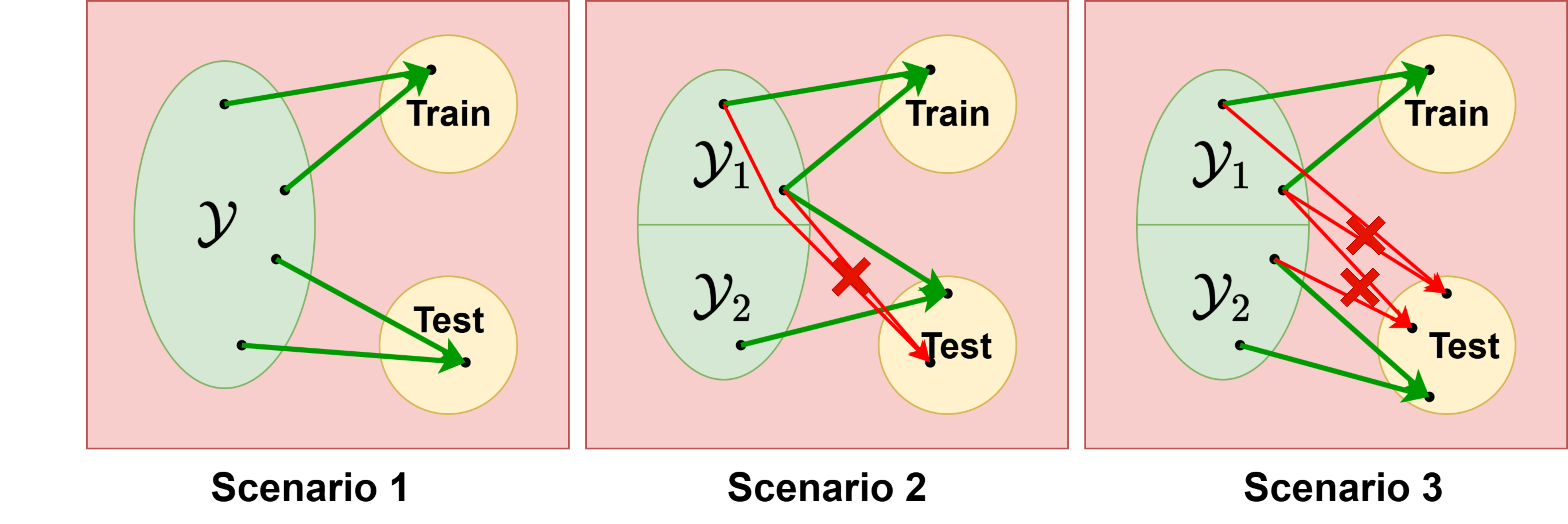}
    \caption{Scenarios in demorphing: Green arrows indicate a valid test morphs whereas red arrows indicate invalid test morphs in each scenario. (Left) Training and testing morphs are created with identities in $\mathcal{Y}$. (Middle) Test morphs are created using one identity in $\mathcal{Y}_1$ and another in $\mathcal{Y}_2$. (Right) Train and test morphs are created using disjoint identities in $\mathcal{Y}_1$ and $\mathcal{Y}_2$, respectively.   }
    \label{fig:scenarios}
\end{figure}

In the literature, demorphing has been explored under different scenarios based on how the training and testing morphs are composed \cite{ref66,dcgan}. 
Consider a set $\mathcal{Y}$ of face images used to create morphs (train and test). The scenarios below are illustrated in Figure \ref{fig:scenarios}.

\begin{enumerate}
    \item \textbf{Same identities in training and testing:} Both train and test morphs are created from face images in $\mathcal{Y}$. However, the same {\em pair} of identities are not used to generate both train and test morphs.

    \item \textbf{Partially unseen identities:} Here, the face set $\mathcal{Y}$ is divided into two disjoint sets with disjoint identities, $\mathcal{Y}_1$ and $\mathcal{Y}_2$. The train morphs are exclusively created from identities in $\mathcal{Y}_1$ whereas test morphs are created using one identity in $\mathcal{Y}_1$ and another in $\mathcal{Y}_2$.
    
    \item \textbf{Completely disjoint identities:} The identities used to create train morphs are completely disjoint from identities used to create the test morphs. In other words, the train morphs are exclusively created from identities in $\mathcal{Y}_1$ whereas test morphs are created exclusively from identities in $\mathcal{Y}_2$.

\end{enumerate}

In this paper, every identity in all of the datasets is represented by a single image. Train morphs are generated using synthetic face images, while test morphs are created exclusively from neutral face images in the FRLL face dataset. Consequently, the terms ``images" and ``identities" are sometimes used interchangeably.

\section{Previous Work}
In \cite{ref51}, the authors propose a GAN-based approach to demorphing inspired by \cite{ref104}. Their method includes a generator and three discriminators. Although the original technique effectively separated natural scene images, it faced a challenge known as morph-replication when applied to demorphing, where the model tended to generate two outputs that closely resembled the morph. This problem arises because of the high degree of local similarity within face images compared to scene images. Moreover, the method assumes that the train and test morphs are created using the same morph technique. Shukla \cite{ref18} introduced a demorphing approach using diffusion models \cite{ref108}, which iteratively adds noise to the morphed image and retrieves the constituent faces during the reverse process. However, this approach assumes that both train and test morphs are derived from the same set of face images, i.e., scenario 1. In \cite{ref66}, the authors propose decomposing the morphed image into multiple unintelligible, privacy-preserving components and then reconstructing the constituent images by weighing and merging these components. This method also presumes shared identities in the train and test morphs. While these techniques achieve varying levels of success, they are generally too restrictive for practical real-world applications. 

In \cite{ref116}, authors propose a reference-based demorphing technique based on diffusion autoencoders. Their method employs pretrained diffusion autoencoders to encode the image into two subspaces: a semantic latent space that captures identity features and a stochastic latent space that retains the remaining stochastic details. The accomplice's face image is restored by decoding the latent code of two subspaces via a conditional denoising diffusion implicit model. FD-GAN \cite{ref17}  is another reference-based method that employs a dual architecture, aiming to reconstruct the first identity's image from the morphed input using the second identity’s image as a reference. To assess the generative model’s effectiveness, it subsequently attempts to recover the second identity using the reconstructed image of the first identity as input.

In this work, we propose a reference-free demorphing technique free of two critical assumptions, i) morph-technique used to create test morphs and ii) the bonafide identities in train and test morphs.

\section{Methodology}
We pose the training of our method as a pixel-wise regression problem in latent space guided by an adversarial loss function and kurtosis regularization. Our method consists of an image compressor, $\mathcal{E}$, a demorphing generator, $\mathcal{G}$, and a discriminator, $\mathcal{D}$. Figure \ref{fig:architecture} shows an overview of our method. Our method first perceptually compresses the input morph using the encoder, $\mathcal{E}$. This step effectively eliminates unimportant high-level visual features that are not critical for demorphing. Another advantage of compressing the input image is the standardization of the representation, i.e., face images are represented in the encoder's latent domain eliminating distractors like background, morph artifacts, lighting, etc. The backbone of our method is based on a conditional-GAN. An image-to-image generator, $\mathcal{G}$, conditioned on the encoded morph images, demorphs the morph in the latent space. A discriminator concatenates the encoded morph with the encoded demorphed outputs and the corresponding ground truth images, and distinguishes between real and synthetic triplets. We optimize the GAN within the latent space of $\mathcal{E}$, which offers two key advantages. (i) Computational efficiency: Since the weights of $\mathcal{E}$ remain frozen, high-resolution face images are represented using lower-dimensional tensors, significantly accelerating both the training and inference processes. (ii) Perceptual Loss: Computing losses in latent space eliminates the need to compare irrelevant features. For instance, the same face image with different backgrounds may exhibit a significantly higher per-pixel loss in the RGB domain, whereas the difference is much smaller in the latent space.  During inference, the decoder, $\mathcal{E}_{dec}$, decodes the demorphed outputs produced by the generator.

\subsection{Face Image Compression} 
 Our face compression model is based on \cite{ref106}, consisting of an autoencoder trained with KL Loss ensuring that the reconstructions are confined to the image manifold by enforcing local realism and preventing the blurriness often caused by relying exclusively on pixel-space losses like $L_2$ or L$_1$. Specifically, given an image $x\in \mathbb{R}^{H\times W\times3}$ in RGB space, the encoder, $\mathcal{E}_{enc}$, compresses the image into a latent representation $z=\mathcal{E}_{enc}(x)\in \mathbb{R}^{h\times w\times c }$. Symmetrically, a decompressor, $\mathcal{E}_{dec}$, reconstructs the image $\Tilde{x}$ from the latent representation such that $\Tilde{x}=\mathcal{E}_{dec}(z)=\mathcal{E}_{dec}(\mathcal{E}_{enc}(x))$. The compressor downscales  the height and width of the image  image by a factor $2^3$ each such that $h=H/8$ and $w=W/8$. In practice, we use a pretrained autoencoder employed in Stable Diffusion \cite{Rombach_2022_CVPR} trained on KL Loss. For an image of size $512\times512\times3$, the autoencoder produces a latent of size $64\times64\times4$. The weights of the autoencoder remain frozen during the entire training process.

\subsection{Demorphing}
We train our demorpher to reduce the distance between the generated demorphed outputs and ground truths in latent space. Unlike prior approaches \cite{ref51, ref18, ref66}, we explicitly order the outputs but randomly swap the ground-truth pairs during training. This allows us to train with varied pair orderings using a standard per-pixel loss (see Figure \ref{fig:ablation}), enabling a more robust generation across both reconstructed images by exposing the model to both possible orders. 

\textbf{Latent conditional GAN:} With our trained compressing network, $\mathcal{E}$, we now perform demorphing in an efficient low-dimensional latent space abstracting away the unnecessary imperceptible details in the RGB pixel space. Performing demorphing in latent space not only makes the process computationally efficient, but also makes it more suitable for the method to focus on important semantic bits of data. The backbone of our method is a conditional GAN \cite{ref47}. The generator, $\mathcal{G}$, and discriminator, $\mathcal{D}$, are conditioned on the morph, $x$. Consider a morph, $x$, created using two face images, $i_1$ and $i_2$, then the conditional GAN loss can be stated as: 
\begin{equation}
\begin{split}
\mathcal{L}_{\text{cGAN}}(\mathcal{G}, \mathcal{D}) = \mathbb{E}_{x, i} [\log \mathcal{D}(x, i)] \\
+ \mathbb{E}_{x, z} [\log (1 - \mathcal{D}(x, \mathcal{G}(x, z)))]
\end{split}
\end{equation}
where, $i=(i_1,i_2)$ represents the outputs. To enforce consistency between the generated and real outputs, we also incorporate an $\mathcal{L}_1$ loss:
\begin{equation}
\mathcal{L}_1 =   \mathbb{E}_{x, i, z} [||i - G(x, z)||_1].
\end{equation}

\noindent\textbf{Morph Replication: } In \cite{ref51}, authors use a GAN to demorph face images which is based on \cite{ref104}. Despite having a separation critic, their method suffered from \textit{morph-replication}, where the model tends to output the morph itself as both of its outputs. This is primarily due to the intrinsic local similarities in face images compared to natural scene images. Indeed, comparing two passport-style face images (typically used in border control, driver's license, etc.) in pixel space would lead to their average being the pairwise distance minimizer and still resembling a face. To alleviate this problem, we introduce another constraint on the prior.  We introduce a kurtosis loss that minimizes the difference in kurtosis between the predicted outputs, $o = (o_1, o_2)$, and the ground truth, $i = (i_1, i_2)$:

\begin{equation}
\mathcal{L}_{\text{kurt}} = \sum_{j=1}^{2} \Big| \text{Kurt}(o_j) - \text{Kurt}(i_j) \Big|
\end{equation}

By aligning the higher-order statistics of the generated and real images, the kurtosis loss helps maintain the structural consistency and realism of the synthesized outputs. This loss is complementary to traditional losses, such as $\mathcal{L}_1$ and perceptual loss, which primarily focus on pixel-wise or feature-wise differences, but may overlook distributional properties. The final objective function incorporates the kurtosis loss alongside adversarial and reconstruction losses:

\begin{equation}
\mathcal{L} = \mathcal{L}_{\text{cGAN}} + \lambda_1 \mathcal{L}_{\text{$1$}} + \lambda_2 \mathcal{L}_{\text{kurt}},
\end{equation}

\noindent where, $\lambda_1$ and $\lambda_2$ are set to $0.5$ each. The inclusion of kurtosis loss enhances the perceptual quality of the generated images by ensuring that the outputs maintain similar statistical characteristics to the real data.


\begin{figure}
    \centering
    \includegraphics[width=\columnwidth]{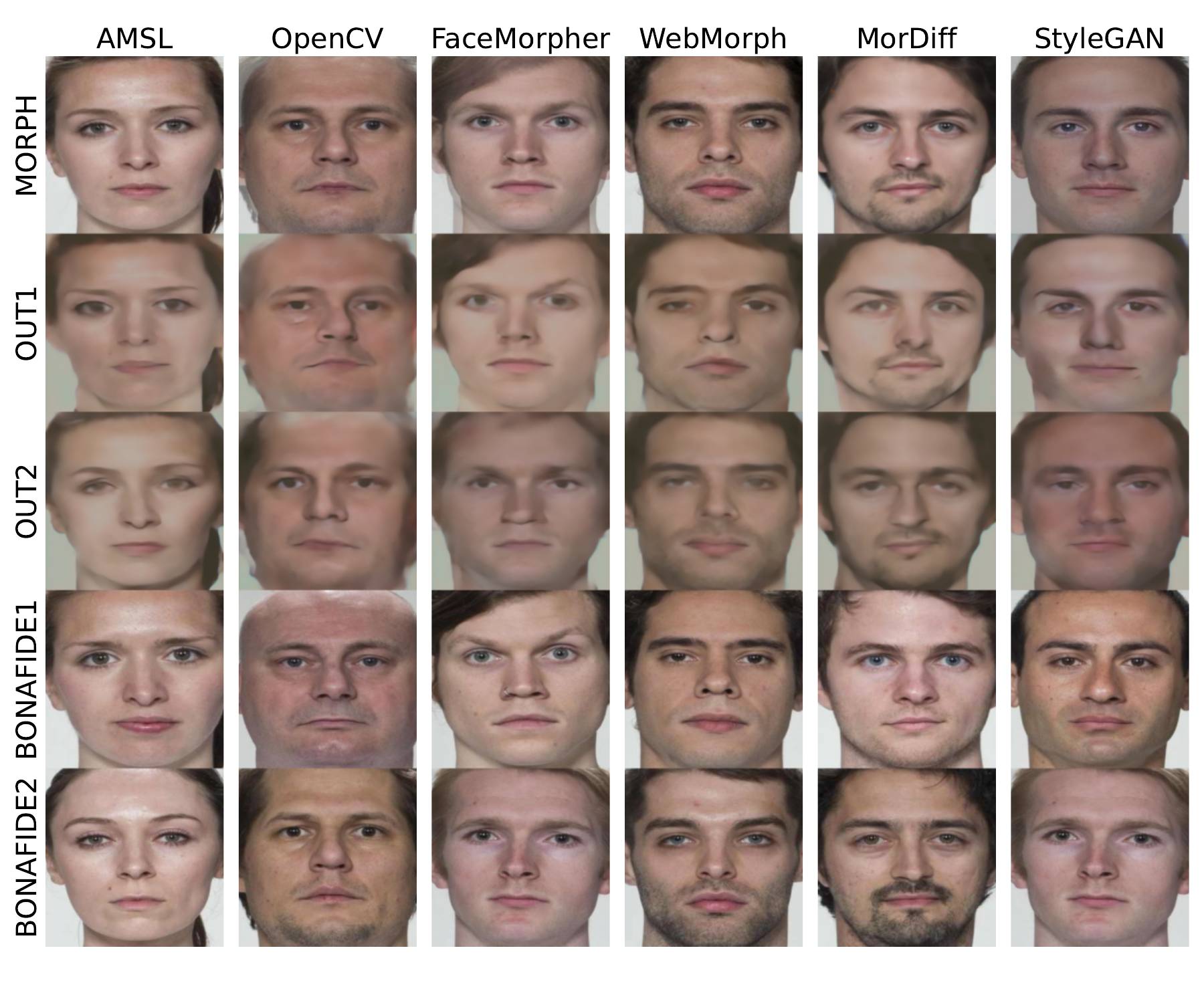}
    \caption{Reference-free demorphing: We test our method on the FRLL-Morph and MorDiff datasets. The model outputs OUT1 and OUT2 in arbitrary order on input MORPH. Our method produces visually distinct outputs i) between OUT1 and OUT2 and ii) between MORPH-OUT1 and MORPH-OUT2.  }
    \label{fig:results}
\end{figure}

\begin{figure}
    \centering
    \includegraphics[width=\columnwidth]{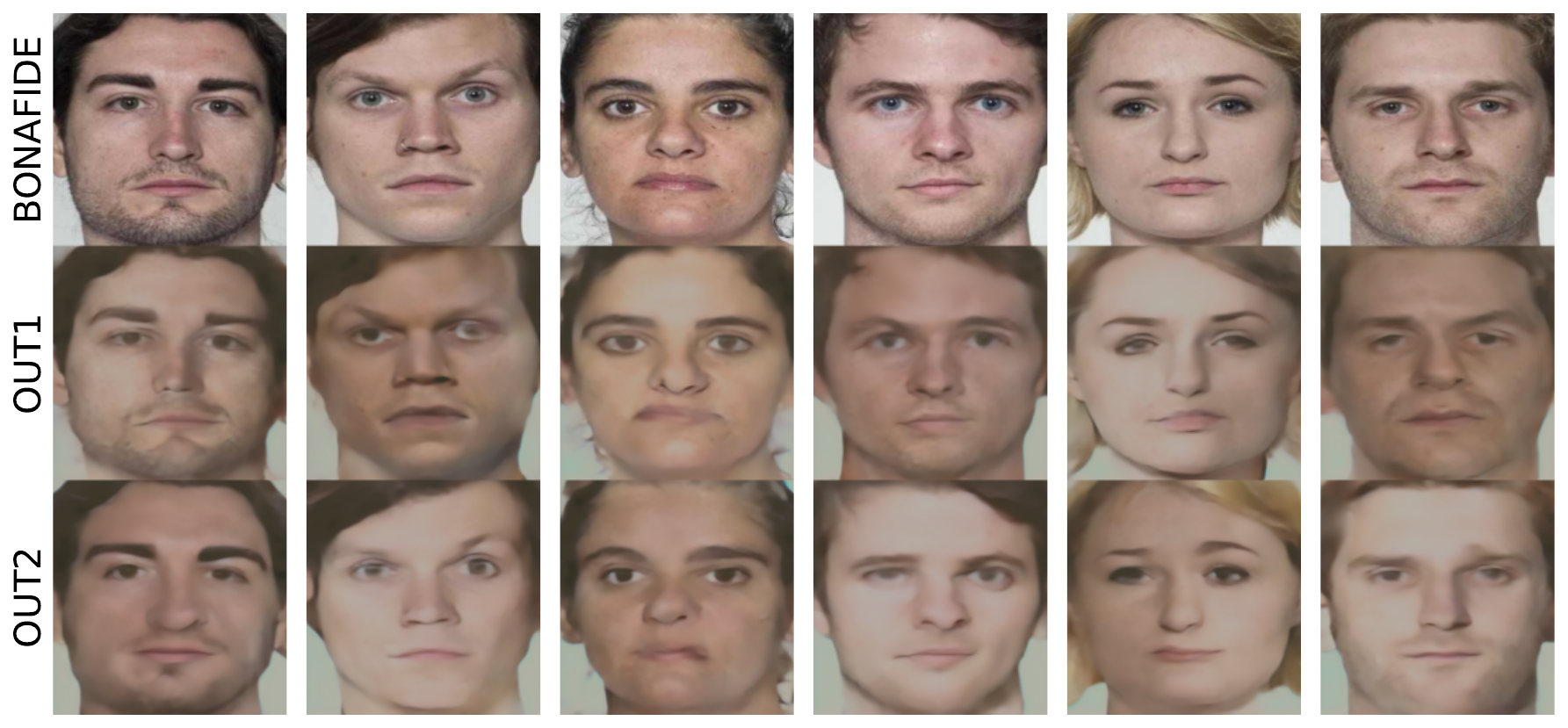}
    \caption{Our method ensures that both outputs maintain the same identity (since input only contains a single identity) when provided with a non-morphed input. }
    \label{fig:enter-label}
\end{figure}

\subsection{Implementation Details}
We use the pretrained autoencoder employed by Stable Diffusion \cite{Rombach_2022_CVPR} trained on KL Loss. The autoencoder consists of four ResNet based downsampling and upsampling blocks each with GroupNorm \cite{gn}, dropout and the SiLU \cite{silu} activation function. Our generator, $\mathcal{G}$, is based on HuggingFace’s conditional UNet implementation, $\texttt{UNet2DConditionModel}$, comprising six ResNet-based downsampling blocks and six corresponding upsampling blocks. Self-attention mechanisms are included in the fifth downsampling block and the second upsampling block. The UNet also requires a timestep input, which we fix to zero in all experiments to neutralize its effect. The discriminator, $\mathcal{D}$, is based on CNN architecture composed of four sequential blocks, each containing a convolutional layer, Instance Normalization \cite{ref76}, and a LeakyReLU activation.  

We evaluate our demorphing method using  AdaFace \cite{ref22} and  ArcFace \cite{ref77} face recognition models. We use cosine similarity to compute the biometric match score. Training is carried out using Adam optimization \cite{ref60} with multi-GPU support via \texttt{accelerate} \cite{ref81}. The training parameters are as follows: number of epochs: 300, learning rate:  $10^{-4}$, dropout rate: 0.1, $\beta_1$: 0.5 and $\beta_2$: 0.999.

\section{Dataset}
\label{dataset}
\textbf{Test datasets: }We test our proposed method on three well-known morph datasets: AMSL \cite{ref64}, FRLL-Morphs \cite{ref65}, and MorDiff \cite{ref9}. The FRLL-Morphs dataset includes morphs generated using four different techniques: OpenCV \cite{ref67}, StyleGAN \cite{ref69}, WebMorph \cite{ref70}, and FaceMorph \cite{ref68}. In all three datasets, the source (i.e., non-morph) images come from the FRLL dataset, which includes 102 identities, each represented by two frontal images — one smiling and one neutral — resulting in a total of 204 face images. The morph counts in each of the datasets are as follows: AMSL: 2,175 morphs; FaceMorpher: 1,222 morphs; StyleGAN: 1,222 morphs; OpenCV: 1,221 morphs; WebMorph: 1,221 morphs; MorDiff: 1,000 morphs. The selection of test datasets includes both conventional landmark-based morph techniques and more recently introduced generative methods.

\begin{table*}[ht]
\centering
\caption{Comparison of our method with the existing state-of-the-art demorphing techniques under a unified protocol. Our method outperforms IPD \cite{ref66}, SDeMorph \cite{ref18} and Face Demorphing \cite{ref51}. We assess our method using established image decomposition IQA metrics (PSNR/SSIM), demorphing metrics (Restoration Accuracy), and biometrically-weighted IQA (BW) \cite{shukla2025metric}. The numbers included for the other methods have been computed by us.}
\label{tab:results}
\resizebox{\linewidth}{!}{
\begin{tabular}{@{}c|c|cccccc|cccccc@{}}
\toprule
\multirow{2}{*}{\textbf{Method}} & \multirow{2}{*}{\textbf{Metric}} & \multicolumn{6}{c|}{\textbf{ArcFace}} & \multicolumn{6}{c}{\textbf{AdaFace}} \\
\cmidrule(lr){3-8} \cmidrule(lr){9-14}
& & AMSL & OpenCV & FMorph & Wmorph & MorDiff & StyleGAN & AMSL & OpenCV & FMorph & Wmorph & MorDiff & StyleGAN \\
\midrule
\multirow{5}{*}{\textbf{Ours}}
& Rest. Acc. @ 10\%FMR & \textbf{99.90\%} & \textbf{99.83\%} & \textbf{100.00\%} & \textbf{100.00\%} & \textbf{100.00\%} & \textbf{38.76\%} & \textbf{98.58\%} & \textbf{99.53\%} & \textbf{99.71\%} & \textbf{100.00\%} & \textbf{100.00\%} & \textbf{50.39\%} \\
& Rest. Acc. @ 1\%FMR  & 99.09\% & 99.30\% & 99.03\% & 99.47\% & 100.00\% & 12.57\% & 96.35\% & 98.58\% & 98.57\% & 98.91\% & 100.00\% & 20.75\% \\
& Rest. Acc. @ 0.1\%FMR  & 96.26\% & 95.83\% & 95.16\% & 96.28\% & 98.63\% & 2.12\% & 91.19\% & 93.69\% & 93.41\% & 94.53\% & 98.83\% & 4.52\% \\
& BW (SSIM) & \textbf{0.44} & \textbf{0.49} & \textbf{0.49} & \textbf{0.46} & \textbf{0.50} & \textbf{0.32} & \textbf{0.35} & \textbf{0.38} & \textbf{0.38} & \textbf{0.36} & \textbf{0.41} & \textbf{0.12} \\
& BW (PSNR) & \textbf{10.12} & \textbf{11.56} & \textbf{11.56} & \textbf{10.56} & \textbf{11.07} & \textbf{7.30} & \textbf{7.92} & \textbf{9.02} & \textbf{9.09} & \textbf{8.22} & \textbf{9.05} & \textbf{2.79} \\

\midrule

\multirow{3}{*}{IPD (2024) \cite{ref66}}
& Restoration Accuracy & 25.69\% & 40.54\% & 37.82\% & 25.61\% & 38.12\% & 16.22\%
& 0.18\% & 1.89\% & 1.43\% & 0.31\% & 3.88\% & 0.00\% \\
& BW (SSIM) & 0.26 & 0.32 & 0.32 & 0.25 & 0.33 & 0.22
& 0.17 & 0.21 & 0.21 & 0.16 & 0.22 & 0.08 \\
& BW (PSNR) & 6.28 & 7.98 & 9.95 & 6.16 & 8.12 & 5.31
& 4.14 & 5.30 & 5.29 & 4.10 & 5.53 & 1.93 \\

\midrule

\multirow{3}{*}{SDeMorph (2023) \cite{ref18}} 
& Restoration Accuracy & 12.56\% & 15.62\% & 13.18\% & 12.80\% & 11.67\% & 0.00\%
& 0.00\% & 0.00\% & 0.00\% & 0.00\% & 0.00\% & 0.00\% \\
& BW (SSIM) & 0.16 & 0.19 & 0.19 & 0.18 & 0.17 & 0.15
& 0.11 & 0.12 & 0.12 & 0.12 & 0.11 & 0.05 \\
& BW (PSNR) & 4.24 & 4.88 & 4.97 & 4.68 & 4.29 & 3.85
& 2.76 & 3.20 & 3.22 & 3.16 & 2.98 & 1.29 \\

\midrule

\multirow{3}{*}{Face Demorphing (2022) \cite{ref51}}
& Restoration Accuracy & 0.45\% & 0.53\% & 0.51\% & 0.50\% & 0.62\% & 0.43\%
& 0.17\% & 0.23\% & 0.17\% & 0.20\% & 0.29\% & 0.00\% \\
& BW (SSIM) & 0.21 & 0.24 & 0.25 & 0.23 & 0.29 & 0.19
& 0.11 & 0.14 & 0.13 & 0.13 & 0.16 & 0.06 \\
& BW (PSNR) & 4.46 & 5.21 & 5.50 & 4.93 & 6.13 & 4.13
& 2.39 & 2.95 & 2.89 & 2.82 & 3.33 & 1.31 \\

\bottomrule
\end{tabular}
}
\end{table*}

\begin{table}[ht]
\centering
\caption{Comparison of demorphing quality using standard image quality assessment metrics: PSNR / SSIM. Higher values indicate better image reconstruction fidelity.}
\label{tab:ssim}
\resizebox{\linewidth}{!}{
\begin{tabular}{@{}c|cccccc@{}}
\toprule
\textbf{Method} & AMSL & OpenCV & FaceMorpher & WebMorph & MorDiff & StyleGAN \\
\midrule

Ours & \textbf{10.81 / 0.47} & \textbf{11.65 / 0.50} & \textbf{11.63 / 0.49} & \textbf{11.17 / 0.49} & \textbf{10.93 / 0.49} & \textbf{10.18 / 0.45} \\
IPD (2024) \cite{ref66} & 9.32 / 0.38 & 10.37 / 0.42 & 10.27 / 0.41 & 9.63 / 0.39 & 9.91 / 0.40 & 9.08 / 0.37 \\
SDeMorph (2023) \cite{ref18} & 8.99 / 0.34 & 9.54 / 0.37 & 9.60 / 0.37 & 9.45 / 0.37 & 8.97 / 0.34 & 8.74 / 0.34 \\
Face Demorphing (2022) \cite{ref51} & 9.68 / 0.46 & 10.35 / 0.48 & 10.44 / 0.47 & 10.20 / 0.48 & 10.13 / 0.47 & 9.51 / 0.45 \\
\bottomrule
\end{tabular}
}
\end{table}


\textbf{Train dataset: }It is widely recognized that generative models require a large amount of training data to capture data distribution accurately. In the context of facial demorphing, a sufficiently large morph dataset is crucial for effective disentanglement. Additionally, the legal and ethical challenges of using/reusing/sharing real biometric data pose a significant challenge. Publicly available datasets are therefore inadequate for training a generalized demorphing technique due to: (i) their limited size (typically around 1.5k morphs) \cite{ref87,ref88,ref8,ref16,ref90,ref91} and (ii) privacy/legal issues \cite{gdpr,gdpr2,ref92,ref93,ref94,ref95}.  To train a high resolution demorphing generative method,we follow the training protocol proposed in \cite{shukla2025metric}. To generate a morph, we  sample two synthetic face images from SMDD train-set which is generated using  StyleGAN2-ADA \cite{stylegan} trained on Flickr-Faces-HQ Dataset (FFHQ) dataset \cite{ffhq}. Choosing random images helps train more generalizable morph attack detection (MAD) systems \cite{ref96}. We create morphed images using the widely adopted \cite{ref88,sarkar2020vulnerabilityanalysisfacemorphing,9093905} OpenCV/dlib morphing algorithm \cite{ref97}, employing Dlib'slandmark detector implementation \cite{ref98} . This landmark-based morphing approach generates morph attacks  that are more effective than those produced by other tools \cite{sarkar2020vulnerabilityanalysisfacemorphing}. With this strategy, we generate 15,000 train and test morphs (from 25K face images in SMDD train and test dataset) each. All images (train and test) are processed using MTCNN \cite{ref71} to detect faces, after which the face regions are cropped. The images are then normalized and resized to a resolution of $512\times512$. Images where faces cannot be detected are discarded. Notably, no further spatial transformations are applied, ensuring that the facial features (such as lips and nose) in both the morphs and the ground-truth constituent images stay aligned throughout training.

\noindent\textbf{Identity Leakage}: Although training with synthetic morphs offers advantages, it is essential to ensure that test identities are not inadvertently replicated during synthetic face generation. In other words, the identities used to create the training morph set should be distinct from those used for generating test morphs. This separation prevents identity leakage and ensures a fair evaluation of the model’s generalization capabilities. We assess identity leakage across 25,000 training face images, each representing a unique identity, and 204 (from 102 identities) test face images from the FRLL dataset. This evaluation is conducted using two face matchers, namely, ArcFace and AdaFace. We observe an average top $n\%$ (0.1,1,5) similarity of 0.22, 0.17, 0.13 between the two sets using the ArcFace matcher and 0.21, 0.16, 0.12 using AdaFace. These relatively low scores suggest that the identities in the test set are unlikely to be present in the training face set, indicating minimal identity leakage.

\begin{table*}[]
    \centering
    \caption{Effects of Separation Priors: We analyze the impact of different separation priors applied during training and find that the combination of kurtosis loss with the standard per-pixel loss yields the best performance among all tested approaches. }
    \resizebox{\linewidth}{!}{
    \begin{tabular}{|c|c||c|c|c|c|c|c|c|}
        \hline
        Metric & Dataset & $\ell_1 + \text{kurt}$ & $\ell_1 \text{ only}$ & $\ell_1 + \text{zhang}$ & $\ell_1 + \text{cross}$ & $\ell_1 + \text{triplet}$ & $\ell_1 \text{ image}$ & trivial \\
        \hline
        \multirow{6}{*}{PSNR} & AMSL  &10.81  &10.75  &10.98  &  9.78& 11.2 &10.0  &  -\\
                            \cline{2-9}
                             &OpenCV  & 11.65 & 11.48 &  11.67&    10.21& 11.83 &10.78 & - \\
                            \cline{2-9}
                             & FaceMorpher & 11.63 & 11.58 &  11.71& 10.25 &  11.99&10.69  &-  \\
                            \cline{2-9}
                             & WebMorph & 11.17 &  11.07&    11.4&  10.12&  11.45& 10.43 & -\\
                            \cline{2-9}
                             & MorDiff &  10.93&   11.03 &  11.33&  9.7&  11.4& 10.16&  -\\
                            \cline{2-9}
                             &  StyleGAN&  10.18&  10.14&  10.37&  9.55&  10.41&  9.53&  -\\
        \hline
          \hline
        \multirow{6}{*}{SSIM} & AMSL  & 0.47 & 0.47 &0.48  &0.39  &  0.49& 0.47 &  1.0\\
                            \cline{2-9}
                             &OpenCV  &  0.5& 0.49 &0.5  &  0.40&  0.51&  0.49&  1.0\\
                            \cline{2-9}
                             & FaceMorpher & 0.49 &0.49  &0.5  &0.39  &0.51  &0.48  & 1.0 \\
                            \cline{2-9}
                             & WebMorph &  0.49&  0.49& 0.5 &  0.41&    0.51 &0.48&  1.0\\
                            \cline{2-9}
                             & MorDiff &  0.49&  0.49&  0.5&  0.39&    0.51&  0.48&1.0\\
                            \cline{2-9}
                             &  StyleGAN&  0.45&0.44  &  0.46&  0.38&    0.47& 0.45&  1.0\\
        \hline
          \hline
        \multirow{6}{*}{ \vtop{\hbox{\strut Restoration Accuracy}\hbox{\strut AdaFace/ArcFace}}} 
                             & AMSL  & 99.89/98.57  & 70.04/53.60        & 71.77/59.20  &  32.72/27.02 & 98.55/97.93 & 99.61/99.33 &  99.11/99.71\\
                            \cline{2-9}
                             &OpenCV  & 99.82/99.52 &  70.55/58.51      & 64.44/54.10 &  36.02/29.17& 99.45/99.05 & 99.81/99.38 &  97.48/97.48\\
                            \cline{2-9}
                             & FaceMorpher & 100/99.71 & 77.10/55.30  & 67.16/54.72&  32.36/27.50& 99.65/98.85 & 100/99.77 & 98.28/98.28 \\
                            \cline{2-9}
                             & WebMorph &  100/100&  72.06/60.16    &73.87/68.12  &  38.94/34.68& 99.46/99.06 &  100/99.84 & 97.5/97.5\\
                            \cline{2-9}
                             & MorDiff &  100/100&  89.74/80.54     &  87.71/85.99&  55.98/34.68&  99.55/99.77&  100/100 &100.00/100.00\\
                            \cline{2-9}
                             &  StyleGAN&  38.56/50.39&  9.12/9.98      &  5.14/6.55&  0.50/0.15&  42.93/48.67&  61.15/67.70 &  97.5/97.5\\
        \hline
          \hline
        \multirow{6}{*}{\vtop{\hbox{\strut BW(PSNR)}\hbox{\strut AdaFace/ArcFace}}} 
                             & AMSL  & 7.92/10.12 & 7.62/10.16 & 3.30/4.82 & 2.69/6.80 & 7.34/9.86 & 6.86/8.40 &  -\\
                            \cline{2-9}
                             &OpenCV  &  9.02/11.56&  8.55/10.98&  3.41/5.14&  2.84/7.07&  7.99/10.23& 7.53/9.26 & - \\
                            \cline{2-9}
                             & FaceMorpher &  9.09/11.56&  8.72/10.86& 3.43/5.38 & 2.85/7.26 & 7.89/10.43 &  7.30/9.38& - \\
                            \cline{2-9}
                             & WebMorph &  8.22/10.56&  7.88/10.36&  3.71/5.08&  2.86/7.07&  7.55/9.81&  6.80/8.42& - \\
                            \cline{2-9}
                             & MorDiff & 9.05/11.07 &  8.66/10.80&  4.6/6.4&  3.07/6.82&  8.57/10.37&  6.82/8.58& - \\
                            \cline{2-9}
                             &  StyleGAN&  2.79/7.3&  2.66/6.99&  1.3/3.8&  1.36/5.62&  2.77/7.09&  2.29/5.83& - \\
        \hline
          \hline
        \multirow{6}{*}{\vtop{\hbox{\strut BW(SSIM)}\hbox{\strut AdaFace/ArcFace}}}
                             & AMSL  & 0.35/0.44  &0.33/0.44  & 0.15/0.21  & 0.11/0.27 & 0.32/0.43 & 0.32/0.40 & - \\
                            \cline{2-9}
                             &OpenCV  &  0.38/0.49&  0.36/0.47&  0.15/0.22&  0.11/0.27&  0.34/0.44& 0.34/0.42  &  -\\
                            \cline{2-9}
                             & FaceMorpher &  0.38/0.49 &  0.37/0.46& 0.15/0.23 & 0.11/0.27 & 0.33/0.44 &  0.33/0.42& - \\
                            \cline{2-9}
                             & WebMorph & 0.36/0.46 & 0.34/0.45 &  0.16/0.22&  0.11/0.28& 0.33/0.43 & 0.31/0.39 & - \\
                            \cline{2-9}
                             & MorDiff &  0.41/0.5&  0.39/0.48&  0.20/0.28&  0.12/0.27&  0.38/0.46& 0.32/0.40 & - \\
                            \cline{2-9}
                             &  StyleGAN&   0.12/0.32 &  0.12/0.30&  0.06/0.17&  0.05/0.22& 0.12/0.31 & 0.11/0.27&- \\
        \hline
        
    \end{tabular}
    }
    
    \label{tab:ablation}
\end{table*}
\begin{figure*}
    \centering
    \includegraphics[width=\linewidth]{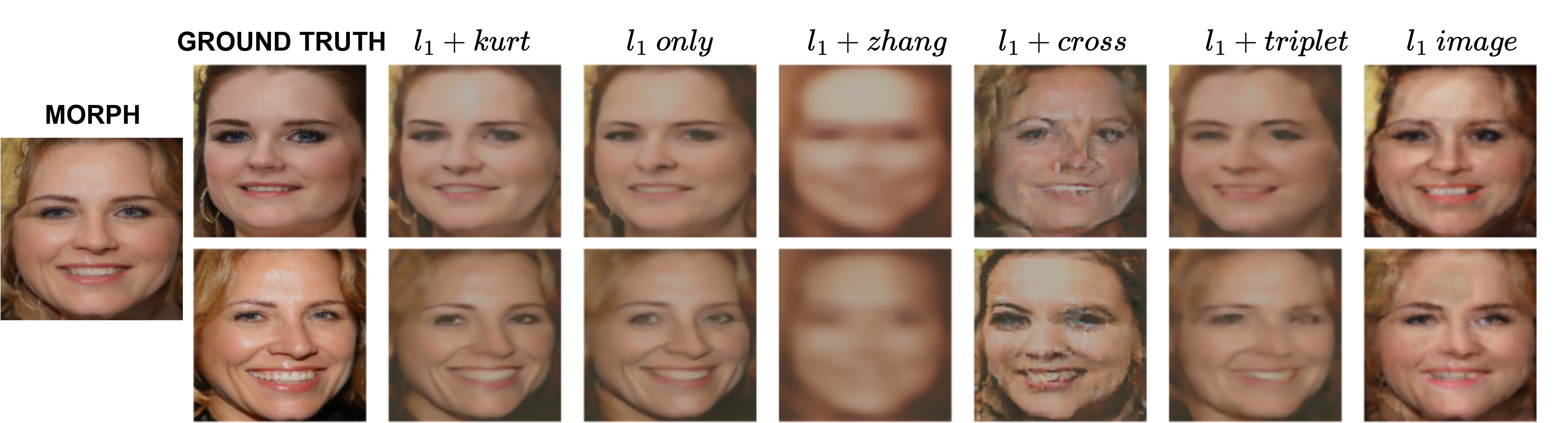}
    \caption{A comparison of the results based on different image separation priors: Zhang’s exclusion loss \cite{ref103}, cross-road loss \cite{ref104} and triplet loss \cite{ref105}. }
    \label{fig:ablation}
\end{figure*}

\section{Evaluation}
\subsection{Existing Metrics}
Reference-free face demorphing is a relatively recent development made possible by advancements in generative learning. There are three main evaluation metrics in the literature to benchmark demorphing methods, namely, True Match Rate at various  False Match Rate thresholds, Restoration Accuracy (RA), and standard Image Quality Assessment metrics such as SSIM and PNSR. 
In \cite{shukla2025metric}, authors argue that TMR@FMR and RA focus only on biometric identity (and not image quality) whereas SSIM/PSNR focus only on quality (and not on biometric utility). Therefore, a comprehensive metric is needed to effectively capture the identities in the face matcher's embedding domain as well as structural quality in pixel-space. This metric, proposed in \cite{shukla2025metric}, is described in the next sub-section.

\begin{figure*}[h]
    \centering
    \includegraphics[width=0.6\linewidth]{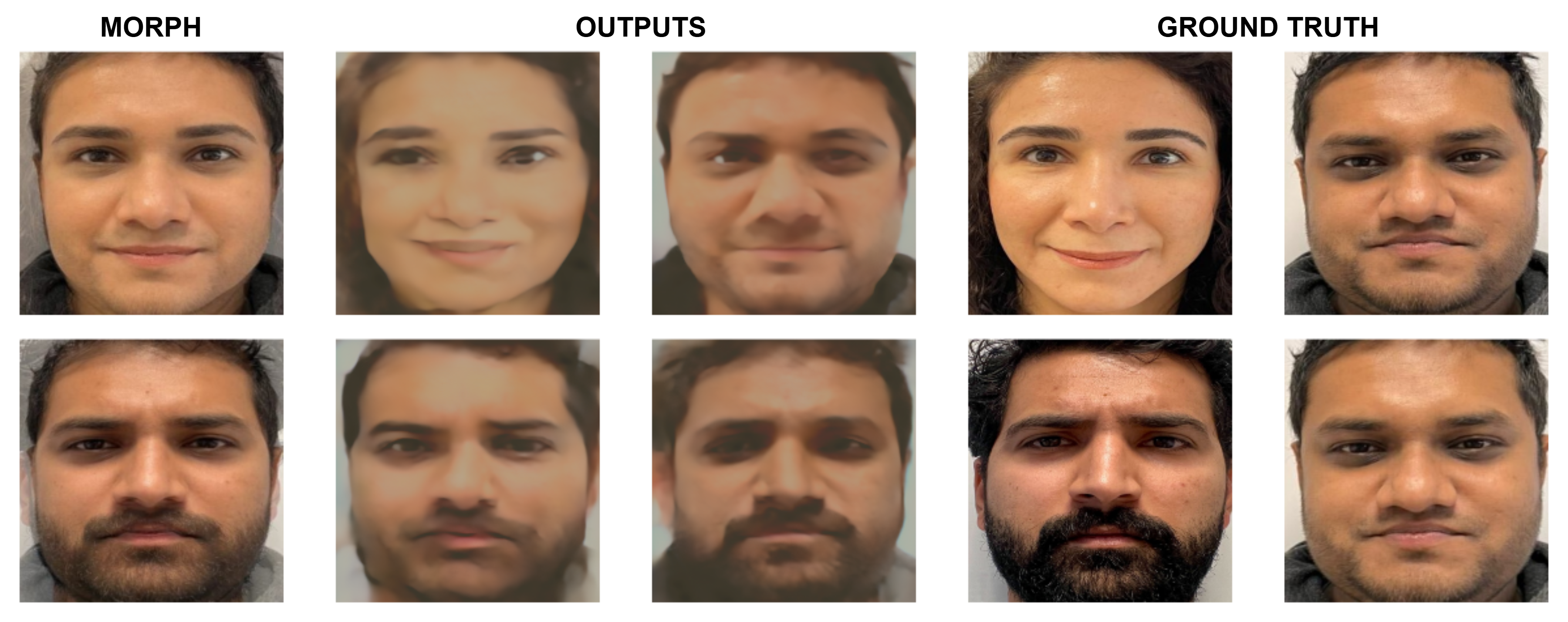}
    \caption{Testing on live subjects: Our method effectively separates the ground truth identities based on gender and facial features. All subjects agreed for publication.  }
    \label{fig:human}
\end{figure*}

\subsection{Biometrically cross-weighted IQA}
Let $\mathcal{X}$ be the face morph set and $\mathcal{Y}$ be the face images used to generate the morphs in $\mathcal{X}$ and $\mathcal{M}:\mathcal{Y}\times \mathcal{Y} \rightarrow\mathcal{X}$ be the morphing operator. A demorpher $\mathcal{DM}:\mathcal{X} \rightarrow \mathcal{Y}\times \mathcal{Y} $ approximates $\mathcal{M}^{-1}$ such that $\mathcal{DM}(x) = (o_1,o_2)$, which may not exist analytically. The goal of $\mathcal{DM}$ is to ensure that \textit{morph replication} does not occur, i.e., $o_1 \neq o_2 \neq x$. In other words, the outputs produced by $\mathcal{DM}$ should look dissimilar  (to the extent they really are), and   the outputs should match the ground truth face images with respect to a biometric matcher $\mathcal{B}$.  The biometrically weighted IQA \cite{shukla2025metric} is defined as
\[
BW(iqa) = \mathbb{E}_{x \in \mathcal{X}} \max \left( 
\begin{aligned}
&\sum_{i\in\{1,2\}} \mathcal{B}(o_i, i_i) \cdot iqa(o_i, i_i), \\
&\sum_{\substack{i\in\{1,2\} \\ j=i\%2+1}} \mathcal{B}(o_i, i_{j}) \cdot iqa(o_i, i_{j}) 
\end{aligned}
\right)
\]  
where, $\%$ is the modulo operator and $iqa \in \{SSIM,PSNR\}$.  BW($iqa$) computes IQA between two possible combinations of output-ground truth pairs and weighs them with the biometric match score produced by a face matcher.

\section{Results}
\label{results}

\subsection{Comparison with State-of-the-Art Methods}
Limited work exists in single-image reference-free demorphing. We compare our methods to  three existing methods.  Table \ref{tab:results} and Table \ref{tab:ssim} show the quantitative comparison of our method against \cite{ref18,ref66}. We compare our method using existing metrics (TMR, Restoration Accuracy, IQA) as well as BW-IQA. Our method significantly outperforms existing methods in the proposed protocol across all metrics and datasets. Visually, our method is able to capture accurate facial features compared to previous methods (see Figure \ref{comparison} and Figure \ref{fig:results}). We also compare our method with \cite{ref51}. On the AMSL dataset, our approach achieves a TMR of 97.77\% using ArcFace, significantly outperforming their result of 70.55\%.

\subsection{Ablation Study}
To evaluate the importance of demorphing in the latent space, we perform the same experiment in the RGB domain, i.e., $\mathcal{E}$ is substituted with an identity function that outputs the input image as it is, keeping other settings unchanged. Table \ref{tab:ablation} shows the evaluation results of the two models and Figure \ref{fig:ablation} shows visual comparisons. Our method in the latent space clearly separates two faces and produces high-resolution images compared to blurry and less distinctive faces when demorphing is done in the pixel space.

\section{Human Study}
\label{sec:human}
To assess the effectiveness of our approach, we perform demorphing on morphs generated from live samples. We gather 17 face images from various subjects (5 males, 3 females), instructing them to pose as if for a driver's license photo, allowing them to decide on wearing or removing accessories such as jewelry or glasses. We created 28 morph images using the protocol described in Section \ref{dataset}, making sure that each pair of identities has at least one morph. We present few samples in Figure \ref{fig:human}. 
Our method generates distinct outputs while preserving a high degree of similarity to the ground truth images. We observe a restoration accuracy of  95.65\% with AdaFace and 91.30\% with ArcFace. We also observe a PSNR of 10.66 and a SSIM value of 0.52. Finally, we evaluate the results using the BW(\textit{iqa}) metrics. We observe a BW(SSIM) of 0.16 and 0.24 on AdaFace and ArcFace, respectively, and a BW(PSNR) of 3.31 and 5.03. These results show the efficacy of the proposed method.


\section{Conclusion}
In this paper, we introduced a novel reference-free  face demorphing framework, Latent Conditional GAN, which operates in a perceptually equivalent latent space to efficiently separate constituent faces from a single morph image. Unlike prior approaches, our method is agnostic to the morphing technique, scales to high-resolution images, and generalizes to unseen morph styles without retraining. By leveraging a latent-space representation, we not only mitigate the morph replication problem but also suppress unnecessary artifacts, ensuring higher facial fidelity in the recovered images. Our method surpasses existing methods by a considerable margin both in terms of analytical performance and visual fidelity. Future work will involve processing morphs generated using more than two images. 

\section{Acknowledgment}

This project was funded by the NSF Center for Identification Technology (Award \#1841517) and DHS CINA (Award \#17STCIN00001-07-01).

{\small
\bibliographystyle{ieee_fullname}
\bibliography{egbib}
}

\end{document}